\documentclass[letterpaper]{article} 
\usepackage{arxiv}  
\usepackage[hyphens]{url}  
\usepackage{graphicx} 
\usepackage{natbib}  
\usepackage{caption} 
\usepackage{algorithm}
\usepackage{algorithmic}

\usepackage{newfloat}
\usepackage{listings}
\DeclareCaptionStyle{ruled}{labelfont=normalfont,labelsep=colon,strut=off} 
\floatstyle{ruled}
\newfloat{listing}{tb}{lst}{}
\floatname{listing}{Listing}

\usepackage{booktabs}

\usepackage{amsmath}
\usepackage{graphicx}
\usepackage{amssymb}
\usepackage{bbding}
\usepackage{booktabs}
\usepackage{tabularx}

\usepackage{colortbl}
\usepackage{xcolor}
\definecolor{mygray}{gray}{.9} 
\definecolor{myred}{rgb}{1,0,0} 
\definecolor{mygreen}{rgb}{0,1,0} 
\definecolor{myblue}{rgb}{0,0,1} 
\definecolor{mycyan}{rgb}{0,1,1} 
\definecolor{mymagenta}{rgb}{1,0,1} 
\definecolor{myyellow}{rgb}{1,1,0} 
\definecolor{myorange}{rgb}{1,0.647,0} 

\usepackage{subcaption}
\usepackage{multirow}
\usepackage{diagbox}
\usepackage{bm}
\usepackage{amsthm}

\definecolor{LightRed}{RGB}{255,182,193} 
\definecolor{LightBlue}{RGB}{173,216,230}

\theoremstyle{plain}

\theoremstyle{definition}

\theoremstyle{remark}

\usepackage{xspace}

\title{HistoLoop: Coarse-to-Fine Spatial Context Integration with a Looped Transformer for Histology-Based Spatial Gene Expression Prediction}

\author{
    Ziang Liu\textsuperscript{\rm 1,2,3},
    Jie Liu\textsuperscript{\rm 1,2,3},
    Shuai Li\textsuperscript{\rm 1,2,3},
    Yigui Feng\textsuperscript{\rm 1,2,3},
    Qinglin Wang\textsuperscript{\rm 1,2,3}\thanks{Corresponding author: Qinglin Wang (wangqinglin@nudt.edu.cn).}
}
\affiliations{
    \textsuperscript{\rm 1}Laboratory of Digitizing Software for Frontier Equipment, National University of Defense Technology\\
    \textsuperscript{\rm 2}National Key Laboratory of Parallel and Distributed Computing, National University of Defense Technology\\
    \textsuperscript{\rm 3}College of Computer Science and Technology, National University of Defense Technology, Changsha 410073, China\\
    
}

\usepackage{bibentry}

\begin{document}

\maketitle

\begin{abstract}
Inferring spatial gene expression from routine H\&E slides requires interpreting local morphology within a tissue organization that spans multiple spatial scales.
Existing multi-scale methods, however, compute regional summaries at each scale only once and then combine them across scales, rather than forming finer-scale summaries from representations that have already absorbed coarser-scale context.
We propose HistoLoop, a looped Transformer that integrates spatial context progressively from coarse to fine.
HistoLoop applies a shared set of modules across recurrent steps at increasingly fine resolutions, and constructs the context at each step from spot states that have already absorbed coarser-scale context.
On HEST-1k, a single model trained jointly on six cancer types achieves a macro-averaged HVG50 Pearson correlation of 0.480, compared with 0.435 for the strongest baseline.
Ablations show that the multi-resolution loop, the coarse-to-fine schedule, and the dependence of each step on updated spot states each contribute to this performance, and clustering analysis indicates that HistoLoop better preserves the multigene structure of tissue domains.
On cancer types held out from training, HistoLoop attains the highest average correlation, although transfer to unseen cancer types remains challenging for all methods.
\end{abstract}
\section{Introduction}

Spatial transcriptomics (ST) profiles gene expression while preserving its
spatial organization within tissue, enabling the characterization of
spatial heterogeneity, tissue microenvironments, and cell--cell
interactions~\cite{stahl2016visualization,longo2021integrating,cang2023commot}.
However, current ST technologies remain costly and experimentally demanding,
which limits their routine application to large cohorts and archival tissue
collections~\cite{moses2022museum}.
In contrast, hematoxylin and eosin (H\&E)-stained whole-slide images are
routinely acquired in clinical pathology and contain rich morphological
signals associated with underlying molecular states~\cite{schmauch2020}.
These complementary properties have motivated histology-based spatial gene
expression prediction, which seeks to reconstruct spatially resolved
transcriptomic profiles directly from standard H\&E images
~\cite{hist2st,bleep,triplex}.

However, the mapping from local histological morphology to molecular
state is inherently context dependent. Morphologically similar tissue
regions may occur in distinct anatomical or pathological environments and
exhibit different transcriptional programs, such that local appearance
alone may provide an ambiguous basis for molecular
prediction~\cite{stem,molf}. Spatial context can help reduce
this ambiguity by relating local morphology to its surrounding tissue
organization~\cite{hist2st,triplex,stflow}. However, if regional context is summarized from spot representations that have not yet been informed by the broader tissue organization, the summary may inherit the very ambiguity it is intended to resolve.
This suggests that finer-scale context may be more reliable when it is constructed after broader tissue organization has been taken into account.

Recent methods have increasingly moved beyond predicting each spot independently from its own image patch by incorporating spatial relationships and broader tissue context. Hist2ST~\cite{hist2st} models both global and neighboring
spatial dependencies through Transformer- and graph-based modules,
TRIPLEX~\cite{triplex} integrates target, neighborhood, and global
tissue information across multiple resolutions, and
STFlow~\cite{stflow} iteratively refines predictions in
gene-expression space using local spatial attention. Most closely related
to our work, HiST~\cite{hist} builds a sparse encoder--decoder
hierarchy that aggregates spot features into progressively coarser tokens
and propagates the resulting context back to finer levels. Together, these
methods establish the importance of nonlocal context for histology-based
spatial gene expression prediction.

The multi-scale designs among them differ in how scales are combined, but they form the regional summaries at each scale once and then reuse them.
In TRIPLEX~\cite{triplex}, target, neighborhood, and global representations are computed separately from the patch features and fused once.
HiST~\cite{hist} does propagate context from coarse to fine in its decoder, but its coarse tokens are aggregated on the fine-to-coarse path with learned, content-independent kernels, and each finer level receives coarser context by upsampling and concatenation with encoder features cached before that context was available.
In both cases, the regional summaries are formed before, and independently of, the coarser-scale context they are later combined with.
A less explored alternative is to re-aggregate the summaries at each finer scale from spot states that have already absorbed coarser-scale context, and to let each spot retrieve context according to its own state.
We ask whether such progressive, state-dependent context construction improves prediction beyond access to multiple spatial scales alone.

To investigate this question, we propose HistoLoop, a looped Transformer that integrates spatial context progressively over recurrent refinement steps.
HistoLoop maintains a persistent state for each spot, initialized from its histological features and spatial coordinates, and assigns each refinement step a spatial resolution following a coarse-to-fine schedule.
At each step, it aggregates the current spot states into context tokens at the assigned resolution, contextualizes these tokens with a spatial Transformer, and retrieves the resulting context back to individual spots to update their states.
The updated states then serve as the input for constructing context at the next, finer resolution, so that finer-scale context is built upon evidence already integrated at coarser scales.
The context construction, spatial Transformer, and retrieval modules are shared across steps, allowing a single set of parameters to learn to integrate spatial context at every resolution, from broad tissue regions to local neighborhoods.
For the state transition, we adopt the memory-based recurrent update of MeSH~\cite{mesh}, which allows spot states to accumulate information across refinement steps.

We evaluate HistoLoop on HEST-1k~\cite{hest} using the HVG50 gene panels of HEST-Bench, training a single model on six cancer types spanning five organs and reserving the remaining four for evaluating generalization to unseen cancer types.
On the six training cancer types, HistoLoop achieves a macro-averaged gene-wise Pearson correlation coefficient (PCC) of 0.480, outperforming the strongest baseline, MoLF~\cite{molf}, at 0.435.
Ablations show that the multi-resolution loop is central to this improvement, with both its coarse-to-fine schedule and the dependence of each step on updated spot states contributing. On the held-out cancer types, HistoLoop attains the highest average PCC, although transfer to unseen cancer types remains challenging for all methods.

Our main contributions are summarized as follows:
\begin{itemize}
    \item We investigate the order and state dependence
    of multi-scale context integration in histology-based
    spatial gene expression prediction, focusing on
    progressive context construction in which finer-scale
    context is formed from spot states that have already
    incorporated coarser-scale information.

    \item We propose HistoLoop, a looped Transformer that progressively refines persistent spot states through coarse-to-fine spatial context construction, with the same modules shared across all resolutions.

    \item We evaluate HistoLoop on HEST-1k under a unified multi-cancer training protocol, where it outperforms recent methods on the training cancer types. Controlled ablations show the contribution of the multi-resolution loop, its coarse-to-fine schedule, and the dependence of each step on updated spot states. A held-out evaluation further shows that transfer to unseen cancer types remains challenging for all compared methods.
\end{itemize}

\section{Related Work}

\subsection{Histology-Based Spatial Gene Expression Prediction}

Histology-based spatial gene expression prediction learns molecular profiles from paired histology and spatial transcriptomics data.
ST-Net~\cite{stnet} established an early image-based prediction framework, while BLEEP~\cite{bleep} and mclSTExp~\cite{mclstexp} explored cross-modal alignment for expression prediction.
DKAN~\cite{dkan} incorporates gene semantic knowledge into this alignment, and AGL~\cite{agl} exploits auxiliary genes to improve target-gene estimation.
Generative approaches provide another direction: Stem~\cite{stem} uses conditional diffusion, STFlow~\cite{stflow} models whole-slide expression through flow matching, and FLAG~\cite{flag} combines latent diffusion with spatial graph encoding and gene foundation model alignment.
MoLF~\cite{molf} further addresses pan-cancer heterogeneity through a mixture-of-experts latent flow.
Broader efforts include HistoPrism~\cite{histoprism}, which emphasizes pan-cancer prediction and pathway-level evaluation, and STPath~\cite{stpath}, which learns transferable expression prediction through large-scale paired pretraining.
These studies advance prediction objectives, biological supervision, and transfer, whereas HistoLoop focuses on how spatial context is integrated across scales.

\subsection{Spatial Context Modeling}

Spatial context modeling relates local morphology to the surrounding tissue organization.
Hist2ST~\cite{hist2st} combines global attention with graph-based neighborhood interactions, while TRIPLEX~\cite{triplex} integrates target, neighborhood, and global features.
HiST~\cite{hist} propagates information through a sparse hierarchical encoder--decoder with spatial coarsening, recovery, and skip connections, and additionally conditions every level on a single slide calibration token that summarizes slide-level acquisition variation.
Slide-level context is also aggregated through self-attention over spot features in HistoPrism~\cite{histoprism} and MoLF~\cite{molf}, and spatial dependencies enter generative models through local spatial attention in STFlow~\cite{stflow} and graph encoding in FLAG~\cite{flag}.
In multi-scale designs such as TRIPLEX and HiST, the regional summaries at each scale are computed once and then combined across scales.
Although the decoder of HiST refines finer levels with upsampled coarse context, its coarse tokens are formed once on the fine-to-coarse path and are never re-aggregated from these refined representations.
HistoLoop instead re-aggregates the regional summaries at each finer resolution from spot states that have already absorbed coarser-scale context, using state-dependent aggregation weights, and lets each spot retrieve context with its own state as the query.

\subsection{Recurrent and Coarse-to-Fine Architectures}

Looped and recursive Transformers reuse a weight-shared block across iterations to increase computational depth without additional parameters~\cite{dehghani2018universal,fan2025looped,saunshi2025reasoning,bae2025relaxed,csordas2024moeut}, and MeSH~\cite{mesh} equips such recursion with memory-based state pathways, whose state transition we adopt.
In dense prediction, shared-weight iterative refinement~\cite{teed2020raft}, including coarse-to-fine processing over image scales~\cite{tao2018srn}, has proven effective on regular image grids.
HistoLoop builds on both lines but differs in what each iteration operates on: each step is assigned its own spatial resolution, and because spots are irregularly distributed, the context tokens at that resolution are re-aggregated from the current spot states rather than taken from the same token sequence or a fixed image pyramid.

\subsection{Pathology Foundation Models}

Pathology foundation models learn transferable histological representations from large-scale data, including patch-level encoders such as UNI~\cite{uni} and CONCH~\cite{conch} and whole-slide models such as GigaPath~\cite{gigapath} and TITAN~\cite{titan}.
Features from such encoders have become the standard spot-level input for spatial gene expression prediction~\cite{hest,stflow,histoprism,molf}.
Following this practice, HistoLoop and all compared methods operate on the same fixed UNI features, so that differences in performance reflect how spatial context is modeled rather than how histological features are encoded.

\section{Method}
\label{sec:method}

\begin{figure*}[t]
    \centering
    \includegraphics[width=0.95\textwidth]{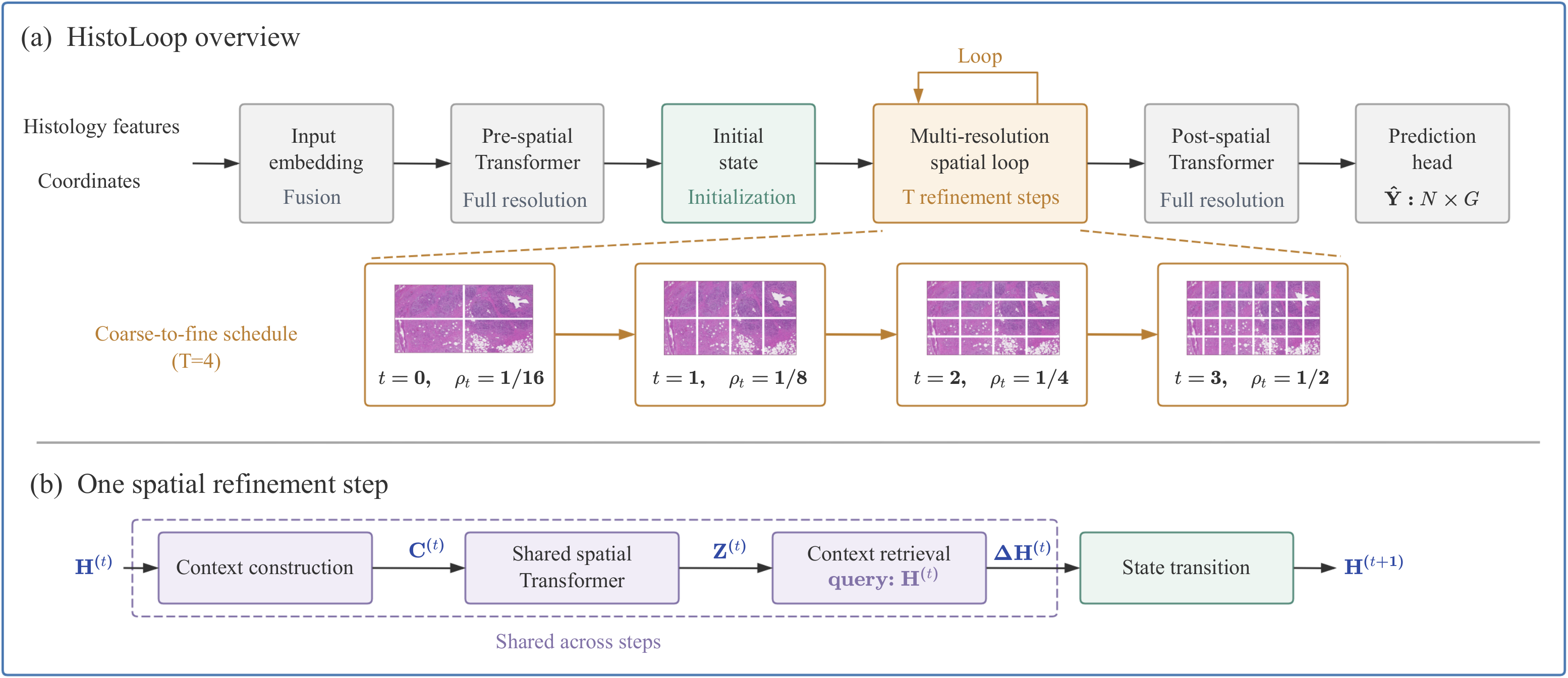}
   
\caption{Overview of HistoLoop. (a) Pretrained histology features and spot coordinates are embedded, contextualized at full resolution, and initialized into persistent spot states. A multi-resolution loop then refines the states over $T$ steps following a coarse-to-fine schedule, in which each step operates at a finer resolution than the previous one and builds on the states it produced. The refined states are decoded into expression profiles $\hat{\mathbf{Y}}$. (b) One refinement step at resolution $\rho_t$. Context tokens are constructed from the current states, contextualized by a spatial Transformer, and retrieved back to each spot using its current state as the query. These three modules are shared across all steps, and the state transition integrates the retrieved update into the spot states.}
\label{fig:overview}
\end{figure*}

\subsection{Overview}
\label{sec:overview}

Given pretrained histological features $\mathbf{X}\in\mathbb{R}^{N\times D}$ of $N$ spots and their spatial coordinates $\mathbf{P}\in\mathbb{R}^{N\times 2}$, HistoLoop predicts the spatial gene expression matrix $\hat{\mathbf{Y}}\in\mathbb{R}^{N\times G}$.
We denote the normalized spot coordinates by $\tilde{\mathbf{P}}=[\tilde{\mathbf{p}}_1;\dots;\tilde{\mathbf{p}}_N]$.
As illustrated in Fig.~\ref{fig:overview}(a), HistoLoop consists of three stages.
The input encoding stage fuses histological and positional information, applies a full-resolution pre-spatial Transformer, and initializes persistent spot states $\mathbf{H}^{(0)}=[\mathbf{h}^{(0)}_1;\dots;\mathbf{h}^{(0)}_N]\in\mathbb{R}^{N\times d}$ together with per-spot $S$-slot memory buffers $\mathbf{M}^{(0)}\in\mathbb{R}^{N\times S\times d}$ (Sec.~\ref{sec:init}).
A recurrent loop then refines the spot states over $T$ steps, producing $\mathbf{H}^{(1)},\dots,\mathbf{H}^{(T)}$, after which a full-resolution post-spatial Transformer and a prediction head produce $\hat{\mathbf{Y}}$ (Sec.~\ref{sec:pred}).

Each refinement step operates at a prescribed spatial resolution.
Following a coarse-to-fine schedule $\boldsymbol{\rho}=(\rho_0,\dots,\rho_{T-1})$, step $t$ summarizes the current spot states into approximately $\lceil\rho_t N\rceil$ context tokens, so that early steps operate on broad tissue regions and later steps on finer spatial structure (Sec.~\ref{sec:schedule}).
Within step $t$, HistoLoop performs four operations:
\begin{equation}
\begin{aligned}
\big(\mathbf{C}^{(t)},\mathbf{P}^{(t)}_{C}\big) &= \mathcal{A}_\psi\big(\mathbf{H}^{(t)},\tilde{\mathbf{P}};\rho_t\big),\\
\mathbf{Z}^{(t)} &= \mathcal{T}_\theta\big(\mathbf{C}^{(t)},\mathbf{P}^{(t)}_{C};\mathcal{G}^{(t)}_{\mathrm{ctx}}\big),\\
\Delta\mathbf{H}^{(t)} &= \mathcal{R}_\phi\big(\mathbf{H}^{(t)},\mathbf{Z}^{(t)};\tilde{\mathbf{P}},\mathbf{P}^{(t)}_{C}\big),\\
\big(\mathbf{H}^{(t+1)},\mathbf{M}^{(t+1)}\big) &= \mathcal{U}_{\mathrm{MeSH}}\big(\Delta\mathbf{H}^{(t)},\mathbf{H}^{(t)},\mathbf{M}^{(t)};t\big),
\end{aligned}
\label{eq:step}
\end{equation}
where $\mathcal{A}_\psi$ denotes state-dependent context construction, which produces context tokens $\mathbf{C}^{(t)}\in\mathbb{R}^{K_t\times d}$ and their weighted coordinates $\mathbf{P}^{(t)}_{C}\in\mathbb{R}^{K_t\times 2}$;
$\mathcal{T}_\theta$ denotes a spatial Transformer operating on a local graph $\mathcal{G}^{(t)}_{\mathrm{ctx}}$ over the context tokens;
$\mathcal{R}_\phi$ denotes context retrieval, which yields a contextual update $\Delta\mathbf{H}^{(t)}\in\mathbb{R}^{N\times d}$;
and $\mathcal{U}_{\mathrm{MeSH}}$ denotes the memory-mediated state transition adopted from MeSH~\cite{mesh}, which updates the spot states and memory buffers.
Because the context at each step is constructed from the current states $\mathbf{H}^{(t)}$, which have already absorbed the context of all coarser resolutions, the context at each finer resolution is built upon coarser-scale context (Sec.~\ref{sec:step}).

The context construction module $\mathcal{A}_\psi$, the spatial Transformer $\mathcal{T}_\theta$, and the retrieval module $\mathcal{R}_\phi$ share their parameters across all refinement steps, whereas the resolution embeddings and the MeSH write--read routers are step-specific.
The input projections, the pre- and post-spatial Transformers, and the prediction head lie outside the loop and have independent parameters.

\subsection{Spot Encoding and State Initialization}
\label{sec:init}

HistoLoop first converts the pretrained features and spot coordinates into full-resolution spot representations and initializes the recurrent states:
\begin{equation}
\begin{aligned}
&\mathbf{E} = f_{\mathrm{img}}(\mathbf{X}) + f_{\mathrm{pos}}(\tilde{\mathbf{P}}),\\
&\mathbf{H}_{\mathrm{pre}} = \mathcal{F}_{\mathrm{pre}}\big(\mathbf{E};\tilde{\mathbf{P}}\big),\\
&\big(\mathbf{H}^{(0)},\mathbf{M}^{(0)}\big) = \mathcal{I}_{\mathrm{MeSH}}\big(\mathbf{E},\mathbf{H}_{\mathrm{pre}}\big),
\end{aligned}
\label{eq:init}
\end{equation}
where $f_{\mathrm{img}}$ and $f_{\mathrm{pos}}$ project the histological features and normalized coordinates into the hidden space, and $\mathcal{F}_{\mathrm{pre}}$ is a pre-spatial Transformer that applies local self-attention on a $k$-nearest-neighbor graph over all spots.
Following MeSH~\cite{mesh}, each spot maintains its own $S$-slot memory buffer.
The initialization $\mathcal{I}_{\mathrm{MeSH}}$ places $\mathbf{E}$ in the first slot of a base memory with the remaining slots set to zero, writes $\mathbf{H}_{\mathrm{pre}}$ into the slots according to a write distribution, and reads $\mathbf{H}^{(0)}$ according to a read distribution, where both distributions are computed from $\mathbf{E}$.
The initial state of each spot is therefore an adaptive combination of its input embedding and its pre-contextualized representation.
The slot-wise initialization is detailed in the supplementary material.

\subsection{Coarse-to-Fine Resolution Schedule}
\label{sec:schedule}

\paragraph{Resolution per step.}
HistoLoop assigns each refinement step a spatial resolution through a coarse-to-fine schedule $\boldsymbol{\rho}=(\rho_0,\dots,\rho_{T-1})$ with $\rho_0<\dots<\rho_{T-1}$, where $\rho_t$ specifies the target number of context tokens relative to the number of spots $N$.
At step $t$, an aspect-ratio-aware grid targeting $\lceil\rho_t N\rceil$ cells is constructed over the bounding box of $\tilde{\mathbf{P}}$, and each spot is assigned to the cell containing its coordinate.
Empty cells are discarded, yielding $K_t$ occupied cells $\{\mathcal{S}^{(t)}_r\}_{r=1}^{K_t}$, each of which produces one context token (Sec.~\ref{sec:step}).
The cell assignment depends only on the spot coordinates and $\rho_t$, whereas the content of each context token depends on the current spot states.
Unless otherwise stated, we use $T=4$ and $\boldsymbol{\rho}=(1/16,1/8,1/4,1/2)$.
The schedule stops at $\rho_{T-1}=1/2$ because $\mathcal{F}_{\mathrm{post}}$, which follows the loop, already models spatial interactions among individual spots and thus serves as the final, full-resolution stage.

\paragraph{A shared Transformer across resolutions.}
The same spatial Transformer $\mathcal{T}_\theta$ processes the context tokens at every resolution, with a step-specific resolution embedding $\mathbf{a}_t$ added to each context token as its only step-specific input.
Attention is restricted to a $k_{\mathrm{ctx}}$-nearest-neighbor graph $\mathcal{G}^{(t)}_{\mathrm{ctx}}$ over the context coordinates, with $k_{\mathrm{ctx}}$ fixed across steps, and relative-position biases are computed from coordinate offsets in the slide-normalized frame.
Because each context token summarizes on average $N/K_t$ spots, a neighborhood of $k_{\mathrm{ctx}}$ tokens covers on average about $k_{\mathrm{ctx}}N/K_t$ spots, and covers all context tokens when $K_t\le k_{\mathrm{ctx}}$.
As $K_t$ increases along the schedule, the spatial extent of each attention neighborhood therefore decreases, from broad tissue regions at early steps to local neighborhoods at later steps.

 \subsection{State-Dependent Refinement Step}
\label{sec:step}

\paragraph{Context construction.}
Given the occupied cells $\{\mathcal{S}^{(t)}_r\}_{r=1}^{K_t}$ at step $t$ (Sec.~\ref{sec:schedule}), the context construction module $\mathcal{A}_\psi$ assigns an importance weight to each spot within a cell and aggregates both the spot states and their coordinates:
\begin{equation}
\begin{aligned}
&\alpha^{(t)}_{ir} = \operatorname{Softmax}_{i\in\mathcal{S}^{(t)}_r}\big(f_\alpha(\mathbf{h}^{(t)}_i,\mathbf{a}_t)\big),\\
&\mathbf{c}^{(t)}_r = \sum_{i\in\mathcal{S}^{(t)}_r}\alpha^{(t)}_{ir}\,\mathbf{h}^{(t)}_i + \mathbf{a}_t,\\
&\mathbf{p}^{(t)}_{C,r} = \sum_{i\in\mathcal{S}^{(t)}_r}\alpha^{(t)}_{ir}\,\tilde{\mathbf{p}}_i,
\end{aligned}
\label{eq:construct}
\end{equation}
where $f_\alpha$ is a two-layer MLP shared across steps, applied to the normalized spot state concatenated with the resolution embedding $\mathbf{a}_t$.
Because the weights $\alpha^{(t)}_{ir}$ depend on the current states, both the content $\mathbf{c}^{(t)}_r$ and the location $\mathbf{p}^{(t)}_{C,r}$ of each context token are determined by $\mathbf{H}^{(t)}$, and so is the graph $\mathcal{G}^{(t)}_{\mathrm{ctx}}$ built over the token locations.

\paragraph{Spatial contextualization.}
The context tokens are processed by the shared spatial Transformer,
\begin{equation}
\mathbf{Z}^{(t)} = \mathcal{T}_\theta\big(\mathbf{C}^{(t)},\mathbf{P}^{(t)}_{C};\mathcal{G}^{(t)}_{\mathrm{ctx}}\big),
\label{eq:contextualize}
\end{equation}
which applies local self-attention over $\mathcal{G}^{(t)}_{\mathrm{ctx}}$ with the relative-position biases described in Sec.~\ref{sec:schedule}.

\paragraph{Context retrieval.}
The contextualized tokens are then mapped back to individual spots.
For each spot $i$, the retrieval module $\mathcal{R}_\phi$ selects the $k_u$ context tokens whose coordinates $\mathbf{p}^{(t)}_{C,r}$ are nearest to $\tilde{\mathbf{p}}_i$, denoted $\mathcal{N}^{(t)}_u(i)$, and computes
\begin{equation}
\Delta\mathbf{h}^{(t)}_i = \mathrm{XAttn}_\phi\Big(\mathbf{h}^{(t)}_i,\big\{\mathbf{z}^{(t)}_r,\mathbf{p}^{(t)}_{C,r}\big\}_{r\in\mathcal{N}^{(t)}_u(i)};\tilde{\mathbf{p}}_i\Big),
\label{eq:retrieve}
\end{equation}
where $\mathrm{XAttn}_\phi$ is multi-head cross-attention with the current state $\mathbf{h}^{(t)}_i$ as the query and the tokens $\mathbf{z}^{(t)}_r$ as keys and values, followed by an output projection.
It uses relative-position biases of the same form as in $\mathcal{T}_\theta$, with separate parameters.
Stacking the updates over all spots gives $\Delta\mathbf{H}^{(t)}\in\mathbb{R}^{N\times d}$.

\paragraph{State transition.}
Following MeSH~\cite{mesh}, the update is integrated into the per-spot memory rather than added directly to the state.
Let $\mathbf{m}^{(t)}_{i,s}$ denote the $s$-th memory slot of spot $i$.
Step-specific routers compute write and read distributions $\mathbf{w}^{(t)}_{\mathrm{write},i},\mathbf{w}^{(t)}_{\mathrm{read},i}\in\mathbb{R}^{S}$ by applying a linear layer and a softmax to the pre-update state $\mathbf{h}^{(t)}_i$, and the transition $\mathcal{U}_{\mathrm{MeSH}}$ computes
\begin{equation}
\begin{aligned}
&\mathbf{m}^{(t+1)}_{i,s} = \mathbf{m}^{(t)}_{i,s} + w^{(t)}_{\mathrm{write},i,s}\,\Delta\mathbf{h}^{(t)}_i,\\
&\mathbf{h}^{(t+1)}_i = \sum_{s=1}^{S} w^{(t)}_{\mathrm{read},i,s}\,\mathbf{m}^{(t+1)}_{i,s}.
\end{aligned}
\label{eq:mesh}
\end{equation}
As analyzed in~\cite{mesh}, this write--read cycle generalizes a residual update: the next state combines an adaptively retrieved summary of the stored states with a gated contribution of the current update.
The memory is maintained independently for each spot, so cross-spot information exchange occurs only through context construction, contextualization, and retrieval.

\subsection{Prediction and Training Objective}
\label{sec:pred}

After $T$ refinement steps, a post-spatial Transformer refines the final states on the same full-resolution spot graph as $\mathcal{F}_{\mathrm{pre}}$, and a prediction head maps them to expression profiles:
\begin{equation}
\begin{aligned}
&\mathbf{H}_{\mathrm{out}} = \mathcal{F}_{\mathrm{post}}\big(\mathbf{H}^{(T)};\tilde{\mathbf{P}}\big),\\
&\hat{\mathbf{Y}} = f_{\mathrm{pred}}\big(\mathbf{H}_{\mathrm{out}}\big).
\end{aligned}
\label{eq:pred}
\end{equation}
Expression targets are mapped to the fixed vocabulary of $G$ genes used in STPath~\cite{stpath}, normalized to 10{,}000 counts per spot using the library size computed over all measured genes, and log1p-transformed.
Because the measured gene panels differ across tissue sections, not every gene in the vocabulary is observed in every section.
Let $\Omega$ denote the set of spot--gene pairs from valid spots and measured genes with finite targets.
We train HistoLoop with the mean squared error over these pairs,
\begin{equation}
\mathcal{L}_{\mathrm{MSE}} = \frac{1}{|\Omega|}\sum_{(i,g)\in\Omega}\big(\hat{y}_{ig}-y_{ig}\big)^2,
\label{eq:loss}
\end{equation}
so that unmeasured entries are excluded rather than treated as zero expression.
The loss covers all measured genes in the vocabulary, while the HVG50 panels are used only for evaluation.
All trainable components are optimized jointly, and the pretrained histological features remain fixed.

\section{Experiments}
\subsection{Experimental Setup}
\label{sec:setup}

\providecommand{\res}[2]{\ensuremath{#1_{\scriptstyle #2}}}
\providecommand{\best}[2]{\ensuremath{\mathbf{#1}_{\scriptstyle #2}}}

\begin{table*}[t]
    \centering
   \caption{HVG50 PCC on the six training cancer types. UNI + MLP predicts each spot independently without slide context. Each entry reports the mean and standard deviation (subscript) over two splits and five random seeds. Macro Avg.\ is the unweighted mean over cancer types, computed for each run. The best mean in each row is shown in bold.}
    \label{tab:main_results}
    \setlength{\tabcolsep}{6pt}
    \renewcommand{\arraystretch}{1.05}
    \begin{tabular}{lcccccc}
        \toprule
        Cancer Type & UNI + MLP & HiST & HistoPrism & MoLF & STFlow & HistoLoop \\
        \midrule
        COAD & \res{0.287}{0.014} & \res{0.329}{0.011} & \res{0.267}{0.025} & \res{0.273}{0.014} & \res{0.343}{0.019} & \best{0.372}{0.008} \\
        IDC  & \res{0.421}{0.016} & \res{0.411}{0.004} & \res{0.356}{0.002} & \res{0.609}{0.007} & \res{0.518}{0.031} & \best{0.653}{0.005} \\
        LUNG & \res{0.338}{0.042} & \res{0.306}{0.029} & \res{0.311}{0.021} & \best{0.513}{0.019} & \res{0.414}{0.024} & \res{0.470}{0.007} \\
        PAAD & \res{0.242}{0.025} & \res{0.322}{0.026} & \res{0.252}{0.020} & \res{0.354}{0.023} & \res{0.405}{0.047} & \best{0.479}{0.011} \\
        READ & \res{0.179}{0.033} & \res{0.263}{0.017} & \res{0.286}{0.006} & \res{0.239}{0.016} & \res{0.266}{0.007} & \best{0.287}{0.006} \\
        SKCM & \res{0.405}{0.041} & \res{0.447}{0.010} & \res{0.386}{0.049} & \best{0.622}{0.009} & \res{0.481}{0.015} & \res{0.621}{0.004} \\
        \midrule
Macro Avg. & \res{0.312}{0.016} & \res{0.346}{0.008} & \res{0.310}{0.013} & \res{0.435}{0.009} & \res{0.405}{0.015} & \best{0.480}{0.004} \\
        \bottomrule
    \end{tabular}
\end{table*}

\paragraph{Data.}
We conduct experiments on HEST-1k~\cite{hest}.
A single model is trained jointly on six HEST-Bench cancer types spanning five organs (COAD, IDC, LUNG, PAAD, READ, and SKCM), and the remaining four cancer types are held out entirely to evaluate generalization to unseen cancer types (Sec.~\ref{sec:generalization}).
For each training cancer type, we follow the two-split protocol of HistoPrism~\cite{histoprism}, which retains the official HEST-Bench test slides of each split and divides the remaining slides into training and validation sets, stratified by cancer type.
Each model takes as input the UNI~\cite{uni} features of all spots in a slide together with their spatial coordinates, and predicts expression over the fixed vocabulary of 38{,}982 genes described in Sec.~\ref{sec:pred}.  Slide counts for each split and cancer type are listed in the supplementary material.

\paragraph{Baselines.}
We compare HistoLoop with four recent methods: HiST~\cite{hist}, HistoPrism~\cite{histoprism}, MoLF~\cite{molf}, and STFlow~\cite{stflow}.
We additionally include UNI + MLP, which predicts each spot from its UNI feature with a two-layer MLP and uses no slide context.
All methods use the same UNI features, spatial coordinates, gene vocabulary, and data splits, and each is trained jointly on the six cancer types.
We keep the architecture, training objective, and model-selection procedure of each baseline as in its official implementation.

\paragraph{Evaluation.}
Following HEST-Bench, we evaluate each cancer type on its predefined panel of 50 highly variable genes (HVG50).
For each test slide, we compute the Pearson correlation coefficient (PCC) between predicted and measured expression across valid spots for each panel gene, and average over genes and test slides to obtain a cancer-level score.
All methods are evaluated against the same processed expression targets.
Each method is trained on both splits with five random seeds (41--45), and we report the mean and standard deviation over the resulting ten runs.
The macro average is computed within each run as the unweighted mean over the six cancer types.

\paragraph{Implementation details.}
HistoLoop uses a hidden dimension of $d=2048$, $T=4$ refinement steps with $\boldsymbol{\rho}=(1/16,1/8,1/4,1/2)$, and three Transformer blocks with eight attention heads in each of $\mathcal{F}_{\mathrm{pre}}$, $\mathcal{T}_\theta$, and $\mathcal{F}_{\mathrm{post}}$.
The full-resolution spot graph uses $k=32$ neighbors, the context graph uses $k_{\mathrm{ctx}}=32$, retrieval uses $k_u=4$ candidates, and each spot has $S=4$ memory slots.
We optimize with AdamW using a learning rate of $5\times10^{-5}$ and a weight decay of 0.05, and select the checkpoint with the lowest validation loss; test slides are not used for model selection.
Further details are provided in the supplementary material.

\subsection{Main Results}
\label{sec:main_results}

Table~\ref{tab:main_results} reports HVG50 PCC on the six training cancer types.
HistoLoop achieves the highest macro-averaged PCC of 0.480, exceeding the strongest baseline, MoLF, by 0.045 and the context-free UNI + MLP reference by 0.168.
It obtains the best mean PCC on four of the six cancer types and ranks second on the remaining two, making it the only method that ranks in the top two on every cancer type.

The improvements vary across cancer types.
The largest gains over the strongest competing method are observed on PAAD (0.074) and IDC (0.044), whereas HistoLoop and HistoPrism perform nearly identically on READ.
MoLF achieves the best result on LUNG and is marginally higher on SKCM, but falls below UNI + MLP on COAD and obtains the lowest score among the context-aware methods on READ, indicating that the relative strength of existing methods depends substantially on the cancer type.
HistoPrism, which is designed primarily for pan-cancer and pathway-level prediction, performs comparably to UNI + MLP on the HVG50 panels.

\begin{table*}[t]
    \centering
    \caption{Ablation study on the six training cancer types. Each entry reports the HVG50 PCC averaged over two splits and five random seeds, with the standard deviation as a subscript. Macro Avg.\ is the unweighted mean over cancer types, computed for each run. The best mean in each column is shown in bold.}
    \label{tab:ablation}
    \setlength{\tabcolsep}{6pt}
    \renewcommand{\arraystretch}{1.05}
    \begin{tabular}{lccccccc}
        \toprule
        Variant & COAD & IDC & LUNG & PAAD & READ & SKCM & Macro Avg. \\
        \midrule
        Full HistoLoop & \best{0.372}{0.008} & \best{0.653}{0.005} & \best{0.470}{0.007} & \res{0.479}{0.011} & \best{0.287}{0.006} & \best{0.621}{0.004} & \best{0.480}{0.004} \\
        \midrule
        \multicolumn{8}{l}{\textit{Multi-resolution loop}} \\
        No loop          & \res{0.325}{0.005} & \res{0.508}{0.008} & \res{0.384}{0.012} & \res{0.379}{0.014} & \res{0.202}{0.009} & \res{0.487}{0.004} & \res{0.381}{0.004} \\
        No loop, deeper  & \res{0.321}{0.006} & \res{0.519}{0.004} & \res{0.388}{0.016} & \res{0.397}{0.009} & \res{0.216}{0.012} & \res{0.473}{0.005} & \res{0.386}{0.006} \\
        \midrule
        \multicolumn{8}{l}{\textit{Resolution schedule}} \\
        Single resolution & \res{0.351}{0.006} & \res{0.617}{0.007} & \res{0.448}{0.013} & \res{0.453}{0.021} & \res{0.272}{0.008} & \res{0.603}{0.003} & \res{0.457}{0.005} \\
        Reverse order     & \res{0.346}{0.011} & \res{0.622}{0.005} & \res{0.446}{0.008} & \res{0.485}{0.016} & \res{0.275}{0.007} & \res{0.603}{0.006} & \res{0.463}{0.003} \\
        \midrule
        \multicolumn{8}{l}{\textit{State dependence}} \\
        Initial-state inputs & \res{0.359}{0.003} & \res{0.628}{0.001} & \res{0.446}{0.010} & \best{0.497}{0.013} & \res{0.274}{0.005} & \res{0.613}{0.006} & \res{0.470}{0.003} \\
        Parallel fusion      & \res{0.338}{0.006} & \res{0.632}{0.004} & \res{0.443}{0.013} & \res{0.491}{0.012} & \res{0.272}{0.009} & \res{0.591}{0.007} & \res{0.461}{0.005} \\
        \bottomrule
    \end{tabular}
\end{table*}
\subsection{Ablation Study}
\label{sec:ablation}

We examine the design of HistoLoop with three groups of controlled variants, summarized in Table~\ref{tab:ablation}.
All variants follow the training and evaluation protocol of Sec.~\ref{sec:setup}, and their detailed definitions are provided in the supplementary material.

\paragraph{Multi-resolution loop.}
\emph{No loop} removes the recurrent loop, including context construction, spatial contextualization, retrieval, and the recurrent state transitions, so that $\mathcal{F}_{\mathrm{post}}$ directly follows $\mathcal{F}_{\mathrm{pre}}$.
\emph{No loop, deeper} further adds four full-resolution Transformer blocks, two in each of $\mathcal{F}_{\mathrm{pre}}$ and $\mathcal{F}_{\mathrm{post}}$, bringing its model size close to that of the full model.
Removing the loop reduces the macro-averaged PCC from 0.480 to 0.381, and the deeper variant reaches only 0.386, which is below the strongest baseline, MoLF (0.435).
The improvement of HistoLoop is therefore not recovered by additional full-resolution computation of comparable size.

\paragraph{Resolution schedule.}
\emph{Single resolution} performs the same four steps at a fixed intermediate resolution of $\rho=1/4$, retaining the step-specific resolution embeddings and routers, and \emph{Reverse order} executes the schedule from fine to coarse, $\boldsymbol{\rho}=(1/2,1/4,1/8,1/16)$.
Both variants keep the number of steps and the shared modules unchanged.
Repeating a single intermediate resolution reduces the macro-averaged PCC to 0.457 and lowers the score on all six cancer types.
Reversing the order yields 0.463 even though it uses the same four resolutions as the full model, indicating that the order in which resolutions are visited, not only the set of resolutions, affects the performance.

\paragraph{State dependence.}
\emph{Initial-state inputs} keeps the full schedule but feeds the initial states $\mathbf{H}^{(0)}$, rather than the current states, to context construction, the retrieval queries, and the MeSH routers at every step; the step updates are still accumulated in the memory and read out after the final step.
\emph{Parallel fusion} also computes the update at each resolution from $\mathbf{H}^{(0)}$, but combines the four updates with learned spot-wise weights instead of the recurrent state transition.
Both variants have access to the same resolutions as the full model, but no step depends on the outputs of preceding steps.
They obtain macro-averaged PCCs of 0.470 and 0.461, respectively, indicating that the benefit of the multi-resolution loop extends beyond access to multiple spatial scales and depends on each step building on the states updated at preceding steps.

\paragraph{Consistency across cancer types.}
The two loop variants and \emph{Single resolution} are lower than the full model on all six cancer types, and the remaining variants are lower on five of the six.
The exception is PAAD, on which \emph{Reverse order} and both state-dependence variants slightly exceed the full model; PAAD is also among the cancer types with the largest run-to-run variation.

\subsection{Tissue Domain Consistency}
\label{sec:clustering}

Gene-wise PCC evaluates each gene independently and does not reveal whether the predicted profiles preserve the multigene structure that distinguishes tissue domains.
Since clustering of expression profiles is commonly used to identify such domains~\cite{zhu2024multi,xiao2025spatially}, we compare clusters derived from the predicted and the measured expression.
For each test slide, we cluster the measured profiles and the profiles predicted by each method separately, applying the same standard pipeline and hyperparameters to all of them: feature scaling, principal component analysis, a $k$-nearest-neighbor graph, and Leiden clustering~\cite{traag2019leiden}.
Spatial coordinates are not used, so that the agreement reflects the structure of the expression profiles themselves.
We measure the agreement between the two partitions with ARI, AMI, NMI, and homogeneity.
Implementation details are provided in the supplementary material.

Table~\ref{tab:clustering} shows that HistoLoop obtains the highest score on all four metrics.
The margin over the strongest baseline, HiST, is small for ARI (0.329 vs.\ 0.319) and larger for AMI (0.526 vs.\ 0.479), NMI (0.530 vs.\ 0.482), and homogeneity (0.611 vs.\ 0.522).
The ranking of the baselines differs from that under gene-wise PCC: HiST, which ranks behind MoLF and STFlow in Table~\ref{tab:main_results}, is the strongest baseline here, indicating that clustering consistency captures an aspect of prediction quality complementary to gene-wise correlation.

\begin{figure}[t]
    \centering
    \includegraphics[width=\columnwidth]{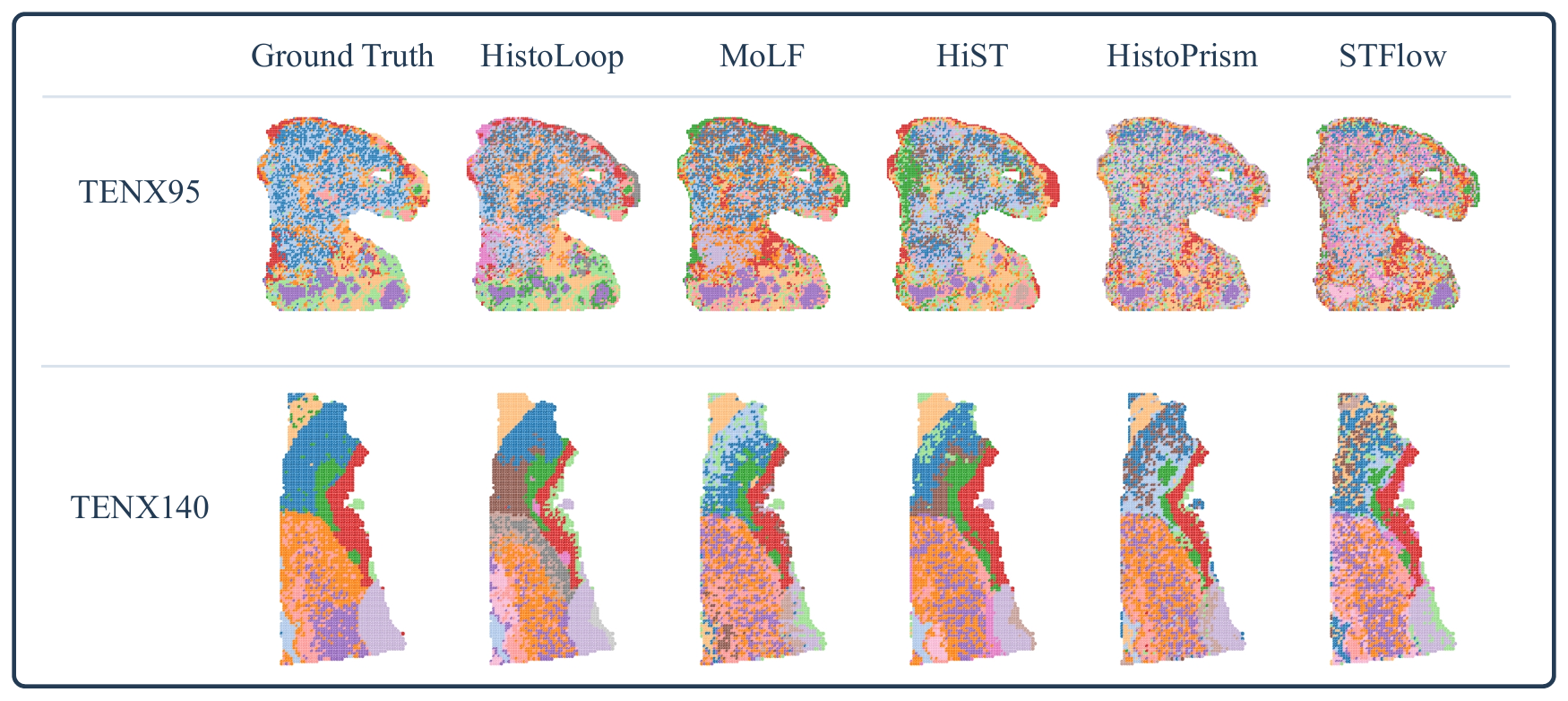}
\caption{Clusters derived from measured and predicted expression on two test slides (seed 42).}
    \label{fig:cluster}
\end{figure}

Figure~\ref{fig:cluster} visualizes the resulting partitions on two test slides, with the clusters of each method displayed at the spot locations.
On both slides, the clusters derived from HistoLoop predictions recover the major regions of the measured partition more coherently, whereas several baselines produce more fragmented or spatially mixed assignments.
These examples are intended for qualitative illustration; the quantitative comparison is based on all test slides in Table~\ref{tab:clustering}.

\begin{table}[t]
    \centering
\caption{Consistency between clusters derived from measured and predicted expression on the test slides of the six training cancer types. Each metric is averaged over all test slides and reported as the mean and standard deviation over two splits and five random seeds. The best mean in each column is shown in bold.}
    \label{tab:clustering}
    \setlength{\tabcolsep}{3pt}
    \resizebox{\columnwidth}{!}{
    \begin{tabular}{@{}lcccc@{}}
        \toprule
        Model & ARI$\uparrow$ & AMI$\uparrow$ & NMI$\uparrow$ & Homo.$\uparrow$ \\
        \midrule
        HiST
        & \res{0.319}{0.011}
        & \res{0.479}{0.005}
        & \res{0.482}{0.006}
        & \res{0.522}{0.006} \\
        HistoPrism
        & \res{0.224}{0.057}
        & \res{0.377}{0.028}
        & \res{0.381}{0.028}
        & \res{0.429}{0.008} \\
        MoLF
        & \res{0.276}{0.003}
        & \res{0.433}{0.026}
        & \res{0.436}{0.027}
        & \res{0.468}{0.028} \\
        STFlow
        & \res{0.211}{0.009}
        & \res{0.360}{0.016}
        & \res{0.364}{0.018}
        & \res{0.406}{0.031} \\
        \midrule
        HistoLoop
        & \ensuremath{\mathbf{0.329}_{\scriptstyle 0.029}}
        & \ensuremath{\mathbf{0.526}_{\scriptstyle 0.002}}
        & \ensuremath{\mathbf{0.530}_{\scriptstyle 0.004}}
        & \ensuremath{\mathbf{0.611}_{\scriptstyle 0.011}} \\
        \bottomrule
    \end{tabular}
    }
\end{table}

\subsection{Generalization to Unseen Cancer Types}
\label{sec:generalization}

Pan-cancer methods such as MoLF~\cite{molf} and HistoPrism~\cite{histoprism} train a single model across cancer types to exploit histology--expression relationships shared across malignancies, but are typically evaluated on cancer types represented in training.
If such relationships are indeed shared, they should also transfer to cancer types absent from training.
We test this by applying the models of Sec.~\ref{sec:main_results} directly to the four held-out cancer types without fine-tuning.
MoLF and HistoPrism condition on the cancer type and learn an unknown category during training in their official implementations, to which we assign the held-out cancer types; the other methods, including HistoLoop, do not take cancer-type information as input.
Since LYMPH\_IDC consists of lymph-node metastases of IDC, a training cancer type~\cite{hoadley2018cell}, we report it separately to examine whether a shared cell of origin facilitates transfer, and average over the remaining three.

\providecommand{\res}[2]{\ensuremath{#1_{\scriptstyle #2}}}
\providecommand{\best}[2]{\ensuremath{\mathbf{#1}_{\scriptstyle #2}}}

\begin{table}[t]
    \centering
\caption{Generalization to cancer types unseen during training. Models trained on the six training cancer types are applied directly to the held-out cancer types. Each entry reports the HVG50 PCC as the mean and standard deviation (subscript) over two splits and five random seeds. Avg.\ is the unweighted mean of the cancer-level means over CCRCC, HCC, and PRAD. LYMPH$^\dagger$ denotes LYMPH\_IDC, which shares its cell of origin with IDC, a training cancer type, and is therefore reported separately. The best mean in each column is shown in bold.}
    \label{tab:generalization}
    \footnotesize
    \setlength{\tabcolsep}{3pt}
    \renewcommand{\arraystretch}{1.05}
    \resizebox{\linewidth}{!}{%
    \begin{tabular}{lcccc@{\hspace{6pt}}c}
        \toprule
        Method & CCRCC & HCC & PRAD & Avg. & LYMPH$^\dagger$ \\
        \midrule
        UNI + MLP  & \res{0.043}{0.015} & \res{0.061}{0.037} & \res{0.109}{0.024} & 0.071 & \res{0.074}{0.022} \\
        HiST       & \res{0.072}{0.005} & \res{0.104}{0.003} & \res{0.121}{0.013} & 0.099 & \res{0.089}{0.002} \\
        HistoPrism & \best{0.097}{0.026} & \res{0.048}{0.018} & \res{0.163}{0.039} & 0.103 & \res{0.104}{0.014} \\
        MoLF       & \res{0.055}{0.019} & \res{0.072}{0.026} & \res{0.176}{0.018} & 0.101 & \res{0.131}{0.031} \\
        STFlow     & \res{0.088}{0.028} & \res{0.043}{0.015} & \best{0.179}{0.030} & 0.103 & \res{0.101}{0.026} \\
        HistoLoop  & \res{0.076}{0.008} & \best{0.131}{0.006} & \res{0.118}{0.014} & \textbf{0.108} & \best{0.134}{0.007} \\
        \bottomrule
    \end{tabular}%
    }
\end{table}

Table~\ref{tab:generalization} shows that the learned relationships transfer only partially.
Even the best held-out result of any method (0.179, STFlow on PRAD) remains well below the lowest score of HistoLoop on the training cancer types (0.287).
Slide context nevertheless remains beneficial, as all context-aware methods exceed UNI + MLP on average (0.099--0.108 vs.\ 0.071).
HistoLoop obtains the highest average PCC over CCRCC, HCC, and PRAD and the best result on HCC, but no method is consistently best: HistoPrism performs best on CCRCC and STFlow on PRAD, where HistoLoop is the lowest among the context-aware methods.
On LYMPH\_IDC, HistoLoop obtains its highest held-out PCC (0.134), closely followed by MoLF (0.131); however, five of the six methods perform best on PRAD instead, so a shared cell of origin does not consistently translate into better transfer in our setting.
\section{Conclusion}
\label{sec:conclusion}

We presented HistoLoop, a looped Transformer that integrates spatial context progressively from coarse to fine for histology-based spatial gene expression prediction.
A shared set of modules is applied across recurrent steps at increasingly fine resolutions, and the context at each step is constructed from spot states that have already absorbed coarser-scale context.
On HEST-1k, HistoLoop achieves the highest macro-averaged HVG50 PCC among the compared methods and ranks in the top two on each of the six training cancer types.
Ablations show that the multi-resolution loop, the coarse-to-fine schedule, and the dependence of each step on updated spot states each contribute to this performance, and clustering analysis indicates that the predicted profiles better preserve the multigene structure that distinguishes tissue domains.

Our evaluation is limited to HEST-1k under the HVG50 protocol.
On the held-out cancer types, HistoLoop obtains the highest average PCC, but all methods perform substantially worse than on the training cancer types, and no method is consistently best.
Improving generalization to cancer types absent from training, and extending the evaluation to additional datasets and sequencing platforms, remain important directions for future work.

\bibliography{main}

\clearpage
\onecolumn
\appendix
\section*{Appendix}

\section{Dataset Statistics}
\label{sec:s_data}

Table~\ref{tab:s_splits} summarizes the two data splits used for the six training cancer types, and Table~\ref{tab:s_test} lists the test slides of each training cancer type. Following HistoPrism~\cite{histoprism}, the test slides are taken from the official HEST-Bench splits, and the remaining slides are divided into training and validation sets stratified by cancer type. Table~\ref{tab:s_heldout} lists the slides of the four held-out cancer types, none of which is used for training or model selection.

\begin{table}[!ht]
    \centering
    \caption{Number of slides in the two data splits for the six training cancer types.}
    \label{tab:s_splits}
    \begin{tabular}{lccc}
        \toprule
        Split & Train & Validation & Test \\
        \midrule
        Split 1 & 275 & 71 & 9 \\
        Split 2 & 277 & 71 & 7 \\
        \bottomrule
    \end{tabular}
\end{table}

\begin{table}[!ht]
    \centering
    \caption{Number of test slides for each training cancer type.}
    \label{tab:s_test}
    \begin{tabular}{lccccccc}
        \toprule
        Split & COAD & IDC & LUNG & PAAD & READ & SKCM & Total \\
        \midrule
        Split 1 & 3 & 1 & 1 & 1 & 2 & 1 & 9 \\
        Split 2 & 1 & 1 & 1 & 1 & 2 & 1 & 7 \\
        \midrule
        Total   & 4 & 2 & 2 & 2 & 4 & 2 & 16 \\
        \bottomrule
    \end{tabular}
\end{table}

\begin{table}[!ht]
    \centering
    \caption{Number of slides for the four held-out cancer types.}
    \label{tab:s_heldout}
    \begin{tabular}{lccccc}
        \toprule
        & CCRCC & HCC & LYMPH\_IDC & PRAD & Total \\
        \midrule
        Slides & 8 & 2 & 2 & 23 & 35 \\
        \bottomrule
    \end{tabular}
\end{table}

\section{Training Details}
\label{sec:s_train}

Table~\ref{tab:s_hparams} lists the training configuration of HistoLoop. Each training sample consists of all spots of one slide.

\begin{table}[!ht]
    \centering
    \caption{Training configuration of HistoLoop.}
    \label{tab:s_hparams}
    \begin{tabular}{ll}
        \toprule
        Setting & Value \\
        \midrule
        Optimizer & AdamW, $\beta=(0.9, 0.95)$ \\
        Peak learning rate & $5\times10^{-5}$ \\
        Weight decay & 0.05 \\
        Learning-rate schedule & linear warmup for 10 epochs, cosine decay to 1\% of the peak \\
        Gradient clipping & global norm 1.0 for the first 100 updates, \\
                          & then per-parameter adaptive clipping (EMA decay 0.99, ratio 1.04) \\
        Maximum epochs & 500 \\
        Model selection & lowest validation loss, early stopping with patience 50 \\
        Random seeds & 41--45 for each split \\
        Hardware & 8 NVIDIA H100 GPUs per run \\
        \bottomrule
    \end{tabular}
\end{table}

\section{Baseline Settings}
\label{sec:s_baselines}

All baselines use the same UNI features, spatial coordinates, 38{,}982-gene vocabulary, and data splits as HistoLoop, and each is trained jointly on the six training cancer types with five random seeds per split. We use the official implementation of each method and keep its architecture, training objective, and model-selection procedure unchanged. For HistoPrism~\cite{histoprism} and MoLF~\cite{molf}, which condition on the cancer type, the held-out cancer types in Sec.~\ref{sec:generalization} are assigned to the unknown category that both methods learn during training in their official implementations. UNI + MLP maps the UNI feature of each spot to its expression profile with a two-layer multilayer perceptron and is trained with the same objective and protocol as HistoLoop.

\section{Ablation Variants}
\label{sec:s_ablation}

All variants in Table~\ref{tab:ablation} share the training protocol of the full model and differ only in the components described below.

\paragraph{No loop.} The recurrent loop, including context construction, spatial contextualization, retrieval, and the recurrent state transitions, is removed, so that $\mathcal{F}_{\mathrm{post}}$ directly follows $\mathcal{F}_{\mathrm{pre}}$.

\paragraph{No loop, deeper.} In addition to removing the loop, two full-resolution Transformer blocks are added to each of $\mathcal{F}_{\mathrm{pre}}$ and $\mathcal{F}_{\mathrm{post}}$, increasing each stage from three to five blocks.

\paragraph{Single resolution.} The loop performs four steps at a fixed resolution of $\rho=1/4$. The step-specific resolution embeddings and MeSH routers are retained.

\paragraph{Reverse order.} The loop executes the schedule $\boldsymbol{\rho}=(1/2,1/4,1/8,1/16)$, and each resolution keeps the resolution embedding associated with it in the full model.

\paragraph{Initial-state inputs.} At every step, context construction (Eq.~\eqref{eq:construct}), the retrieval queries (Eq.~\eqref{eq:retrieve}), and the MeSH routers (Eq.~\eqref{eq:mesh}) take the initial states $\mathbf{H}^{(0)}$ as input instead of the current states $\mathbf{H}^{(t)}$. The updates of all steps are still written to the per-spot memory, and the state is read from the memory after the final step. Since no step uses the states read out at preceding steps, these intermediate readouts do not affect the prediction.

\paragraph{Parallel fusion.} For each resolution $\rho_s$ of the schedule, a contextual update $\Delta\mathbf{H}_s$ is computed from $\mathbf{H}^{(0)}$ with the shared context construction, spatial Transformer, and retrieval modules. The four updates are combined with spot-wise weights,
\begin{equation}
\mathbf{h}_i = \mathrm{Norm}\Big(\mathbf{h}^{(0)}_i + \sum_{s} \beta_{i,s}\, \Delta\mathbf{h}_{i,s}\Big), \qquad
\beta_{i,s} = \operatorname{Softmax}_{s}\big(g([\mathbf{h}^{(0)}_i;\Delta\mathbf{h}_{i,s};\mathbf{a}_s])\big),
\label{eq:s_parallel}
\end{equation}
where $g$ is a two-layer MLP with hidden dimension $d/4$ shared across resolutions. This fusion replaces the recurrent state transition, and the fused states are passed to $\mathcal{F}_{\mathrm{post}}$.

\section{State Initialization}
\label{sec:s_init}

Following MeSH~\cite{mesh}, the initialization $\mathcal{I}_{\mathrm{MeSH}}$ in Eq.~\eqref{eq:init} constructs a base memory for each spot $i$ by placing its input embedding $\mathbf{e}_i$ in the first slot,
\begin{equation}
\mathbf{m}^{\mathrm{base}}_{i,1} = \mathbf{e}_i, \qquad \mathbf{m}^{\mathrm{base}}_{i,s} = \mathbf{0} \;\; (s>1).
\label{eq:s_base}
\end{equation}
Write and read distributions are computed from $\mathbf{e}_i$ by linear layers followed by a softmax over slots, and the pre-contextualized representation $\mathbf{h}^{\mathrm{pre}}_i$ is written into the memory before the initial state is read out:
\begin{equation}
\mathbf{m}^{(0)}_{i,s} = \mathbf{m}^{\mathrm{base}}_{i,s} + w^{\mathrm{init}}_{\mathrm{write},i,s}\,\mathbf{h}^{\mathrm{pre}}_i, \qquad
\mathbf{h}^{(0)}_i = \sum_{s=1}^{S} w^{\mathrm{init}}_{\mathrm{read},i,s}\,\mathbf{m}^{(0)}_{i,s}.
\label{eq:s_init}
\end{equation}
Unlike the transitional step of MeSH, which computes these distributions from the prelude output, we compute them from the input embeddings, consistent with computing the routers at each recurrent step from the incoming state. We use $S=4$ slots.

\section{Clustering Protocol}
\label{sec:s_cluster}

For each test slide, the measured expression is normalized to log1p(CP10k), and the predictions of all methods are on the same scale. Each expression matrix is restricted to genes with nonzero variance across spots, from which the 5{,}000 most variable genes are retained. The selected genes are standardized (clipped at 10), reduced to 50 principal components, and connected in a 15-nearest-neighbor graph, on which Leiden clustering~\cite{traag2019leiden} is performed with a resolution of 0.5. The same procedure and hyperparameters are applied to the measured expression and to the predictions of every method, and spatial coordinates are not used. ARI, AMI, NMI, and homogeneity are averaged over all test slides and then over the ten runs.


\end{document}